# Robotic Handling of Compliant Food Objects by Robust Learning from Demonstration

Ekrem Misimi, Alexander Olofsson, Aleksander Eilertsen, Elling Ruud Øye, John Reidar Mathiassen

*Abstract*—The robotic handling of compliant and deformable food raw materials, characterized by high biological variation, complex geometrical 3D shapes, and mechanical structures and texture, is currently in huge demand in the ocean space, agricultural, and food industries. Many tasks in these industries are performed manually by human operators who, due to the laborious and tedious nature of their tasks, exhibit high variability in execution, with variable outcomes. The introduction of robotic automation for most complex processing tasks has been challenging due to current robot learning policies. A more consistent learning policy involving skilled operators is desired. In this paper, we address the problem of robot learning when presented with inconsistent demonstrations. To this end, we propose a robust learning policy based on Learning from Demonstration (*LfD*) for robotic grasping of food compliant objects. The approach uses a merging of RGB-D images and tactile data in order to estimate the necessary pose of the gripper, gripper finger configuration and forces exerted on the object in order to achieve effective robot handling. During *LfD* training, the gripper pose, finger configurations and tactile values for the fingers, as well as RGB-D images are saved. We present an *LfD* learning policy that automatically removes inconsistent demonstrations, and estimates the teacher's intended policy. The performance of our approach is validated and demonstrated for fragile and compliant food objects with complex 3D shapes. The proposed approach has a vast range of potential applications in the aforementioned industry sectors.

SUPPLEMENTARY MATERIAL

For supplementary video see: https://youtu.be/cafFC9HgaFI

## I. INTRODUCTION

Contact processing, or the interaction of a robot with objects that require manipulation, is one of the greatest challenges facing robotics today. Most approaches to robotic grasping and manipulation are purely vision-based. They either require a 3D model of the object, or attempt to build such models by full 3D reconstruction utilizing vision [1]. In many areas, knowledge of the 3D model of an object can be sufficient for a robot to achieve optimal grasping [2]. However, a lack of detailed information about the object's surface, and mechanical and textural properties, can affect grasping accuracy, causing handling to be much more challenging.

These challenges become even greater if the objects requiring manipulation are compliant, because there is an additional requirement not to degrade their quality. These issues are even greater if the objects also require dynamic dexterous manipulation, such as when they are moving. This is highlighted in particular when it comes to handling and grasping movements during harvesting, post-harvesting and production line processing operations across entire value chains involving fragile and compliant raw materials of oceanic and agricultural origin. Compliancy challenges both the classical approaches based on rigid-body assumptions [3], as well as reliance on purely vision-based approaches to robotic grasping and handling. Recently, the ocean space, agriculture and food sectors have increased their interest in flexible, robot-based, automation systems, suitable for small-scale production volumes and able to adapt to natural biological variation in food raw materials [4]. For example, the food processing industry is characterised by many manual tasks and complex processing operations that rely heavily on the dexterity of a human hand. Optimal grasping, manipulation and imitation of the complex manual dexterity of skilled human operators by robots is thus prerequisite for greater uptake and utilisation of robotic automation by the ocean space, agricultural and food industries [4, 5].

In some of the most complex processing operations, such as manually-based meat or fish harvesting, cutting or trimming, the skilled human operator uses a number of visual and tactile senses, and combines these with a task description and previous experience [4, 5]. For example, during cutting, a skilled operator will utilise his visual and tactile senses to adapt his efforts and make adjustments in order to execute essential changes in his hand position and the force he exerts to complete the task. This is an important skill for a robotic system to learn to be able to perform the same task with similar efficiency. This example illustrates the way in which humans

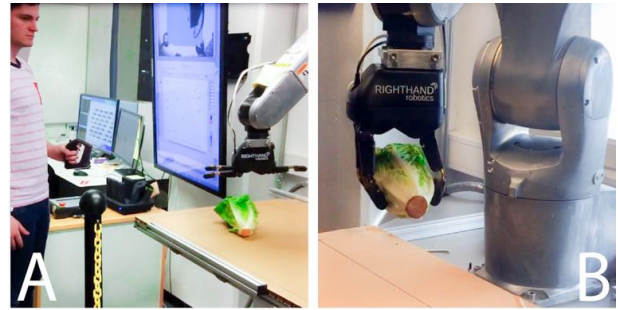

Figure 1. a) *Learning from Demonstration*: Teacher demonstrating grasps of compliant objects by remote control of the 6-DOF robot arm and gripper; b) *Autonomous grasping*: after learning the grasping skills from the teacher, the robot autonomously grasps compliant objects.

E. Misimi, A. Olofsson, A. Eilertsen, E.R. Øye, and J. R. Mathiassen are with SINTEF Ocean, Trondheim, NO 7465 Norway (phone: +47 98222467; e-mail: ekrem.misimi@sintef.no).

approach solving grasping and handling problems by using multi-modal perception combined with both their visual and tactile senses [6]. Both are essential to optimal grasping and handling, and are particularly important when grasping fragile and compliant objects such as food raw materials. Combining two senses is essential to a grasping action that is both gentle and efficient. A further challenge facing the food processing and similar sectors, where skilled operators perform complex operations, is the inconsistency of the operators' performance. In a scenario in which a robot has to be taught by humans to perform such complex operations, the accuracy of robotic manipulation will vary depending on the teacher and the high levels of variability involved in the teaching approach. Our approach here is motivated by the need to develop a consistent learning policy that is independent of teacher variance. This in turn alleviates the burden of accuracy on the teachers.

The key contributions of this work are: a) an *LfD* learning policy that automatically discards the inconsistent demonstrations by human teachers, addressing the problem of learning when presented with inconsistent human demonstrations; b) the integration of visual (RGB-D) and tactile data into space-action pairs for the grasping of compliant objects using the aforementioned *LfD* learning policy, thus enabling the automatic rejection of inconsistent data and the effective autonomous grasping of fragile and deformable food objects. Based on visual input (RGB-D images), we estimate the pose (6 DOF) and gripper configuration for grasping and, based on tactile data, we estimate the forces necessary to achieve the optimal grasping of previously unseen compliant objects. Both visual input and tactile data are gathered during the training stage implemented using an *LfD* approach that is in turn facilitated by teleoperation control of the robotic grasping action using STEM controllers. This paper demonstrates the validity of the approach by illustrating how a robot is taught to grasp a fragile and compliant food object, in this case lettuces.

## II. RELATED WORK

Robotic grasping and handling of objects has been studied extensively and a wide variety of approaches can be found in the literature. Most of the research reported in the field is dedicated to robotic grasping and handling of rigid bodies [1, 7, 17], but recently research is also being focused on deformable objects [3, 8], and also compliant objects of different origin with application domains including ocean, agriculture and food processing industry [4, 5, 9, 10]. Most of the research reported on robotic grasping and manipulation is also based on developing models that estimate the grasping of objects solely based on the visual input data [7, 10, 11]. While vision enables localization in 3D of the objects of interest, the benefit of using tactile data is in accurate perception of the compliancy of the objects to be manipulated. This is particularly important when manipulating food objects whose quality cannot be degraded once the contact with the robot is established and during the interaction [4, 5]. In addition, robotic handling approaches based purely on vision are limited because food raw material comes with a high biological variation, and although they may have the same visual properties, they can exhibit varying material and mechanical properties based on how they are treated during the production in the value chain. Although the use of tactile sensors has been demonstrated for different robotic manipulation applications [12, 13], including grasping of fragile compliant objects [14] and food objects [9], the adoption of tactile sensors for robotic manipulation applications has been slow. This has been due to the difficulty to both integrate tactile sensors in the standard schemes but also due to the challenges to capture and model the tactile readings [3, 15]. Even scarcer are the cases of multi-modal perception combinations of visual and tactile information, mainly focused on robotic grasping of some standard rigid objects [15]. For example, Motamedi et al. [16] explore the use of haptic tactile and visual feedback during object-manipulation tasks assisted by human participants to shed light on how human approach to grasping combines the tactile and visual feedback. In both [15, 16], although for rigid objects, it is demonstrated how incorporating tactile sensing may improve the robotic grasping of objects.

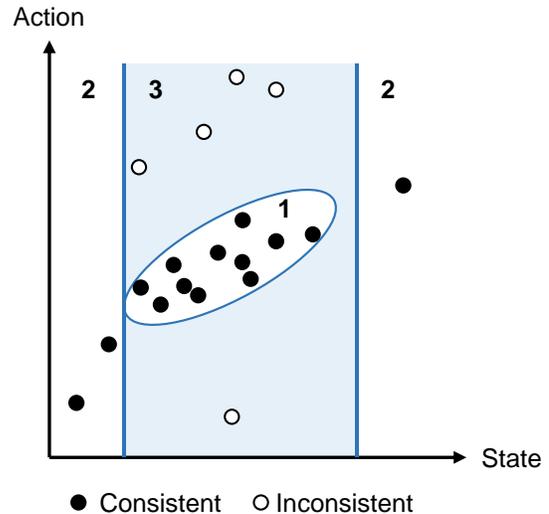

Figure 2. Illustration of consistent and inconsistent demonstrations, and the three relevant regions in the demonstration (state-action) space – (1) state-action inliers, (2) state outliers and (3) state-action outliers that are not state outliers. Demonstrations in region (1) and (2) are consistent.

One aspect of our work is inspired by the existing literature and the lack of multi-modal visual and tactile perception for robotic manipulation of highly fragile and compliant food raw materials, and especially regarding the use of RGB-D image data with tactile data, differing from the [15] in that they use pure 2D images in combination with tactile data for grasping of rigid objects. Our approach combines visual and tactile sensing and feeding these inputs to a robust learning policy, based on *LfD*. This policy enables not only to estimate the 6DOF grasping pose, by means of visual input, but also the necessary gripper configuration and forces exerted to the food object to accomplish handling without quality degradation.

One particular learning policy to use when endowing robots the autonomy to perform manipulation tasks on compliant food objects is the Learning from Demonstration (*LfD*). *LfD* policy is learned from examples, or demonstrators, executed by a human teacher, and these examples constitute the state-action pairs that are recorded during the teacher's demonstrations [18]. This learning policy has been used to enable robust learning using regularized and maximum-margin methods [19], but has also been applied to robotic grasping [20], more specifically to robot grasp pose estimation [21] and grasping based on input from 3D or depth images [11, 22]. A characteristic of *LfD* is that it relies on successful demonstration of the desired task(s) from a human teacher, and that researchers usually manually discard the poor demonstrations. Recently there has been reported research work investigating the possibility to learn from so-called poor demonstrations [23, 24], namely to also learn what not to imitate. In the context of these previous studies and in the context of the outlined challenge regarding the need to alleviate the accuracy burden placed on the human teacher during the demonstration, our work differs from related work in the aspect that it presents a *LfD* algorithm that is able to automatically discard the inconsistent demonstration examples, prior to learning the teacher's *intended* policy. This enables the learner (robot) to act more consistently and with less variability in task performance compared to the teacher.

## III. PROBLEM STATEMENT

The goal of this work is to learn a robust policy, in a tele-operated *LfD* context, which is applicable to tactile-sensitive grasping of compliant objects based on visual input. Grasping tasks have a few unique challenges, including the possibly high-dimensional visual state space, and difficulty of achieving accurate teleoperation. Many visual features may be needed in order to represent the object state in sufficient detail to be able to predict a grasp. The number of training examples may be many times the number of features to create an accurate prediction model. The possibly tedious nature of one or more human teachers providing hundreds of grasp demonstration examples, may lead to some inconsistencies and errors made by teacher(s). Additionally, the action of using teleoperation for grasping may in itself create additional variance and inconsistencies in the demonstration examples. This motivates us to explicitly develop a robust policy learning algorithm that derives a policy only from consistent demonstrations. We consider a demonstration set $D = (S, A) = \{d_i\}_{i=1}^{n}$ of $n$ demonstrations $d_i = (s_i, a_i) \in \mathcal{D}$, each consisting of a state $s_i \in \mathcal{S}$ and an action $a_i \in \mathcal{A}$, where $\mathcal{D} = \mathcal{S} \times \mathcal{A}$ is the demonstration space, $\mathcal{S}$ is the state space and $\mathcal{A}$ is the action space.

Using the mapping function approach to LfD, we seek to learn from the demonstrations $D$ a parameterized policy of the form $\pi_\theta : \mathcal{S} \mapsto \mathcal{A}$, where $\theta$ is the parameter vector associated with the policy. We let the teacher's *executed* policy be denoted by $\tilde{\pi} : \mathcal{S} \mapsto \mathcal{A}$ and *intended* policy be denoted by $\pi^* : \mathcal{S} \mapsto \mathcal{A}$. The executed policy may contain unintended demonstration examples, inaccuracies and inconsistencies, and may be seen as a corrupted version of the teacher's intended policy. Let $D^* = (S^*, A^*)$ be the subset of the demonstration examples for which the teacher followed its intended policy. Despite following its intended policy for these demonstration examples, there may still be some noise (e.g. jitter in the hand of the teacher or measurement noise). Therefore, although $\pi^*$ exists, we can only measure noisy realizations $\hat{\pi}^*(s) \in A^*, s \in S^*$ of this intended policy. From these realizations, we find an optimal parameterized policy $\pi_\theta^*$ that estimates the teacher's intended policy $\pi^*$. This is done by minimizing the expected loss function $\ell : \mathcal{A} \times \mathcal{A} \mapsto \mathbb{R}$ over the probability distribution induced by the states $s \in S^*$. The optimal parameters are found by solving

$$\min_\theta E_{s \in S^*}[\ell(\pi_\theta(s), \hat{\pi}^*(s))]. \qquad (1)$$

For our use, we assume that the state distribution is stationary, and that the parameters therefore can be found using supervised learning in offline mode under that assumption.

## IV. ROBUST POLICY LEARNING

### A. Discovering Demonstrations Consistent with the Teacher's Intended Policy

In this section we will describe a method for discovering demonstrations consistent with the teacher's intended policy $\pi^*$, from the executed policy $\tilde{\pi}$ given by the demonstrations $D$. This requires the assumption that the teacher actually has an intended policy. We let a consistent policy imply that, for a given state, the same action or set of actions will be taken. Besides noise, the executed policy may have two main types of deviations from this consistency: 1) deviation in intention, 2) deviation in execution. The first deviation type simply implies that the teacher did a mistake in what he intended – e.g. intending to pull the nail out of the board instead of intending to hammer it into the board. The second deviation implies a mistake in execution – e.g. intending to hit the nail, but missing it and hitting the board instead. If we can discover the demonstration examples representing these types of deviations from consistency, then we can possibly learn a consistent policy from the remaining demonstrations.

Formally, we define a demonstration $d_i = (s_i, a_i)$ to be consistent if its probability in demonstration space satisfies $P_D(d_i) > \rho_D$, *or* its probability in state space satisfies $P_S(s_i) < \rho_S$, where $P_D$ and $P_S$ are probability densities induced by the demonstrations $D$ and states $S$, respectively. Here, $\rho_D, \rho_S \in [0,1]$ are respective probability thresholds. Assuming we have access to the exact empirical probability density functions $P_D$ and $P_S$, the consistent demonstrations are then found exactly as

$$D_{exact}^* = \{d_i \in D : P_D(d_i) > \rho_D \lor P_S(s_i) < \rho_S\}. \quad (2)$$

We can gain an intuitive understanding of this from Figure 2. The consistent demonstrations are found in two regions in the demonstration (state-action) space. The first region corresponds to the demonstrations that are in a high-probability region of the *demonstration space*. The second region corresponds to those demonstrations that are of low probability in the *state* space, and therefore where one cannot conclude whether they are consistent or not. One may argue that these outliers in state space could simply be removed from the data set, however in some settings one may wish to keep as many demonstrations as possible due to the small size of the data set. However, the biggest reason for keeping the state outliers is to allow for a policy that is valid in the entire sampled *state* space, also in extreme states. In some cases, there may be unwanted consequences if extreme states cannot be handled, such as e.g. strong wind gusts and actuator failure when flying an autonomous drone. In these cases, it is better to have a potentially inconsistent policy than no policy at all. Keeping state outliers also enables us to use larger values of $\rho_D$, to remove as many inconsistent examples as possible in more populated regions of the demonstration space.

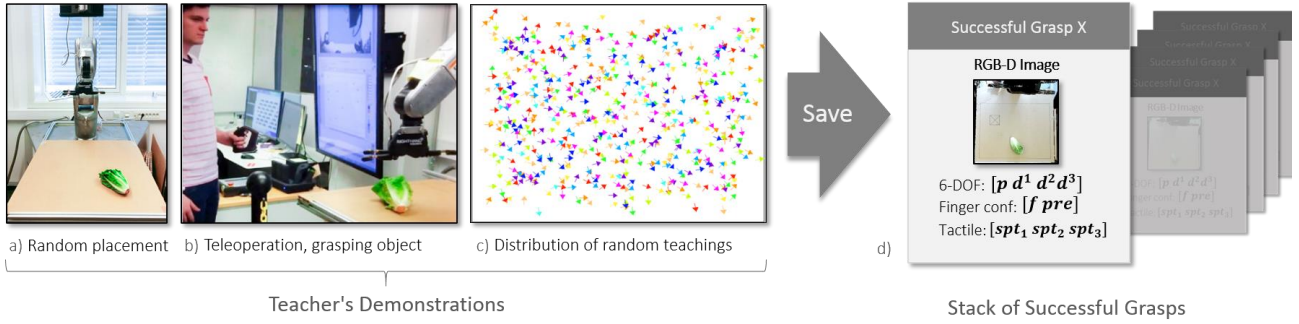

Figure 3. Generation of the dataset for the training phase. a) The setup with RGB-D sensor, robot arm with ReFlex TakkTile gripper and the lettuce as an example of fragile compliant food object placed in the grasping area. b) Human teacher teleoperating the robot arm with STEM controllers to demonstrate the grasps. c) During the generation of the dataset, 525 training examples were generated by random placement of different lettuces. Lettuces of different sizes and shapes were placed on random position and orientation in the grasping area. d) For the demonstrated grasps, the following data were saved constituting the state space: RGB-D image, 6DOF of the gripper hand, finger configuration, and tactile significant thresholds for the gripper fingers.

In practice, it is difficult to compute (2), since this requires the exact empirical probability density functions $P_D$ and $P_S$. Even obtaining estimates of these density functions may require a number of samples (i.e. demonstrations) that scale exponentially in the number of dimensions [25], which in our case is the number of dimensions in the demonstration and state spaces. An alternative, to using an exact empirical probability density function and a given probability threshold, is to estimate point membership decision functions that approximate a given quantile level set, such as the one-class support vector machine (SVM) method proposed in [26, 27]. This approach has previously been successfully used in a *LfD* context [23] to determine point membership in an approximate quantile level set in the *state* space. We use the one-class SVM method to compute two decision functions $g_{D,v_D}: \mathcal{D} \mapsto \{-1,1\}$ and $g_{S,v_S}: \mathcal{S} \mapsto \{-1,1\}$, using their respective one-class SVM optimization parameters $v_D, v_S \in [0,1]$ that specify the given quantile and thus the fraction of 'outliers' that result in a negative sign on the decision function. Based on this, we compute an approximate set of consistent demonstrations by

$$D^* = \{\mathbf{d}_i \in D : g_{D,v_D}(\mathbf{d}_i) > 0 \lor g_{S,v_S}(\mathbf{s}_i) < 0\}. \quad (3)$$

The one-class SVM is applicable to high-dimensional spaces with nonlinear decision boundaries, by using a feature map $\Phi : \mathcal{X} \mapsto F$, from some set $\mathcal{X}$ (in our case, $\mathcal{D}$ or $S$) into a dot product space $F$, such that the dot product in that space can be computed using a simple kernel [28, 29] $k(\mathbf{x}, \mathbf{y}) = \Phi(\mathbf{x}) \cdot \Phi(\mathbf{y})$, such as the radial basis function kernel

$$k(\mathbf{x}, \mathbf{y}) = e^{-\gamma \|\mathbf{x}-\mathbf{y}\|_2^2}.$$

Given a set of $l$ feature vectors in $\mathcal{X}$, the solution of the dual problem of the one-class SVM for a given parameter $v$, provides us with the coefficient vector $\boldsymbol{\alpha}$ and bias $\rho$ that, together with the kernel, defined the decision function

$$g_v(\mathbf{x}) = \text{sgn}\left(\sum_i \alpha_i k(\mathbf{x}_i, \mathbf{x}) - \rho\right),$$

where the coefficients $\boldsymbol{\alpha}$ are found by solving the optimization problem

$$\min_{\boldsymbol{\alpha}} \frac{1}{2} \sum_{i,j} \alpha_i \alpha_j k(\mathbf{x}_i, \mathbf{x}_j)$$

$$s.t.\ 0 \leq \alpha_i \leq \frac{1}{vl}, \sum_i \alpha_i = 1,$$

and where the bias is recovered by computing $\rho = \sum_j \alpha_j k(\mathbf{x}_i, \mathbf{x}_j)$, where $i$ is any $i$ for which the corresponding coefficient satisfies $0 < \alpha_i < \frac{1}{vl}$, i.e. it is non-zero and is not equal to the upper bound.

The optimal value of $\gamma$ is found by a one-dimensional cross-validated search for the value that minimizes the number of predicted outliers.

B. *Policy Learning using $\varepsilon$–insensitive Support Vector Regression*

Using the approximate set of consistent demonstrations $D^* = (S^*, A^*)$ found in (3), we will now present a method for learning the policy $\boldsymbol{\pi}_{\boldsymbol{\theta}}^*$ by solving (1). For this purpose, we use $\varepsilon$–insensitive support vector regression (SVR), which has previously been used to predict grasps from 3D shape [30]. As a regression method, SVR has similar robustness to outliers and good generalization ability [28], as is also the case with SVM classification [27, 28].

The loss function for which $\varepsilon$–insensitive SVR is a minimization algorithm, is the $\varepsilon$–insensitive loss function for scalar action arguments

$$l_\varepsilon(a, a^*) = \begin{cases} 0, & |a - a^*| \leq \varepsilon \\ |a - a^*| & \text{otherwise.} \end{cases}$$

In the case of $d$-dimensional vectors of scalar actions, we define the loss function

$$l_{\boldsymbol{\varepsilon}}(\mathbf{a}, \mathbf{a}^*) = \sum_{k=1}^{d} l_{\varepsilon_k}(a_k, a_k^*) \quad (4)$$

We let $\pi_{\boldsymbol{\theta}_k}(\mathbf{s})$ denote the policy for determining the scalar action $a_k$. The parameter vector $\boldsymbol{\theta}_k$ contains the parameters that define the soft-margin radial basis function kernel SVR model [31], which we write as

$$\boldsymbol{\theta}_k = [\boldsymbol{\alpha} \quad b \quad \mathbf{S} \quad \gamma]_k,$$

where $\alpha$ are the $n_{sv,k}$ non-zero coefficients, $b$ is the bias, $\mathbf{S}$ contains the $n_{sv,k}$ support vectors, and $\gamma$ is the radial basis function kernel scale parameter. An additional parameter, $C$, is the box constraint used in SVR. The optimal values of $C$ and $\gamma$ are found by performing a cross-validated grid-search [32] that minimizes the expected loss function on the validation set.

---

**Algorithm 1** Learning teacher's intended policy

**Input:** Demonstration set $D$, one-class SVM parameters $v_D, v_S$, and SVR parameter $\varepsilon$.
1. Compute demonstration set $D^*$ consistent with teacher's intended policy, using equation (3) and one-class SVM.
2. Find teacher's intended policy by minimizing equation (1) using $D^*$ and the loss function $l_\varepsilon(\mathbf{a}, \mathbf{a}^*)$ from equation (4) using SVR.

**Output:** Parameters $\boldsymbol{\theta}_k = [\boldsymbol{\alpha} \quad b \quad \mathbf{S} \quad \gamma]_k$ defining the policy $\pi_{\boldsymbol{\theta}_k}(\mathbf{s})$, defined by equation (5).

---

Using the SVR parameters, we obtain the following expression for the policy

$$\pi_{\boldsymbol{\theta}_k}(\mathbf{s}) = \sum_{i=1}^{n_{sv,k}} \alpha_{i,k} e^{-\gamma_k \|\mathbf{S}_{i,k} - \mathbf{s}\|_2^2} - b_k, \quad (5)$$

where $\mathbf{S}_{i,k}$ denotes the support vector indexed by $i$ in parameter vector $\boldsymbol{\theta}_k$. The resulting *LfD*-based learning policy that automatically discards inconsistent teacher's demonstrations is given in Algorithm 1.

## V. IMPLEMENTATION DETAILS

### A. Robot platform and Software System

Our experiments were carried out on a 6-DOF Denso VS 087 robot arm mounted on a steel platform. The robot arm was controlled with a software written in LabVIEW, using functions that control the robot arm through Cartesian position coordinates, and two vectors describing the orientation of two of the axes of the end effector relative to the robot's base coordinates. For developing a learning policy based on *LfD*, the teleoperation of the robot arm was done using a hand-held Sixense STEM motion tracker controllers. The STEM-system consisted of five trackers and one base station. Each controller is equipped with a joystick and several buttons. This allowed for wireless feedback of not only controllers' pose but also enables trigging of various events. The pose of the motion trackers was given relative to the position and orientation of the base station, so it was necessary to keep the base station in a fixed orientation relative to the robot arm. For gripping, we used the ReFlex TakkTile hand from Right Hand Robotics. The hand has 3 fingers with a joint feedback, one fixed, and two fingers that can rotate to alter their pose. Each finger is able to bend at a finger joint and is equipped with 9 tactile sensors. The hand is by default controlled in ROS environment. To control the hand via LabVIEW, a simple Python hand control script was run on a Raspberry Pi 3 (RPi3) and then LabVIEW was connected to RPi3. RPi3 is a credit-sized mini-computer with great capabilities similar to a PC. Our Python scripts work by importing rospy, a pure Python client for ROS, which is known to work with example code supplied alongside with the ReFlex hand. After each run, the position and orientation values of the robot arm were saved. In addition, we saved the different tactile values for each finger when grasping the compliant object, and the angles of each joint (Fig. 3). The tactile values were used when grasping to calculate how hard to close the hand and what forces should be exerted to the object for a successful grasping. A Kinect v2 camera was mounted on a rod perpendicularly looking at the robot arm and the grasping scene where the compliant objects were placed for grasping (Fig. 3). A full HD (High Definition) colour image with $1920 \times 1080$ resolution, IR image and depth image of $512 \times 424$ resolution was registered for each grasping example (Fig. 3). These images were taken several times for each example, once at the beginning, saving the initial position of the object. Another set of images was saved once the hand has been moved to the proper position and the object was grasped, just before the object was to be released and finally once the arm has been moved back to the initial position. The last image is significant because it is used to determine if the grasping was a success. These images were used for feature extraction (Figure 3) while developing the learning policy for autonomous grasping based on *LfD*.

### B. Dataset and Data Augmentation

Since *LfD* policy is learned from examples, or demonstrators, executed by a human teacher, these examples were generated as described in Fig. 3. To generate the examples each object to be grasp was randomly placed into the grasping area (Fig. 3a) before the grasping was demonstrated with teleoperation (Fig. 3b). 20 different lettuces were used to generate 525 examples evenly spread over the grasping area (Fig. 3c). For each example, a set of data is saved constituting of RGB-D image, gripper finger configuration, and tactile values for the 3 gripper fingers. A data augmentation procedure was used over the generated examples to increase the dataset. Data augmentation was performed by translation of the coordinate space and flipping the gripper for 180 degrees, by randomly perturbing states $x_a$, $y_a$, and $z_a$ with zero mean Gaussian noise with a standard deviation of 25 mm. Training of the policy was done using a split of 75% of the data, and validation was done on the remaining 25%. Validation and demonstration of our *LfD* approach was done using a batch of green lettuces. From the grasping and handling point of view, lettuce is an irregular, fragile, compliant object of variable shape and size. Lettuce is highly fragile product that requires adaptability when it comes to tactile sensing and exertion of forces during grasping. From the visual point of view, lettuce is 3D model-free object and has a biological variability that makes it a relevant and interesting object to test our *LfD* approach based on RGB-D images and tactile perception. The lettuce blade has a compliancy and variability in structure that changes from one end to the other, making the lettuce as a representative and challenging compliant object to work with both from the visual and tactile point of view.

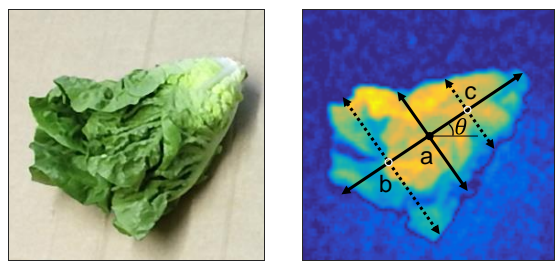

Figure 4. Close-up of the colour (left) and depth (right) images of a lettuce, with overlaid extracted visual features on the depth image.

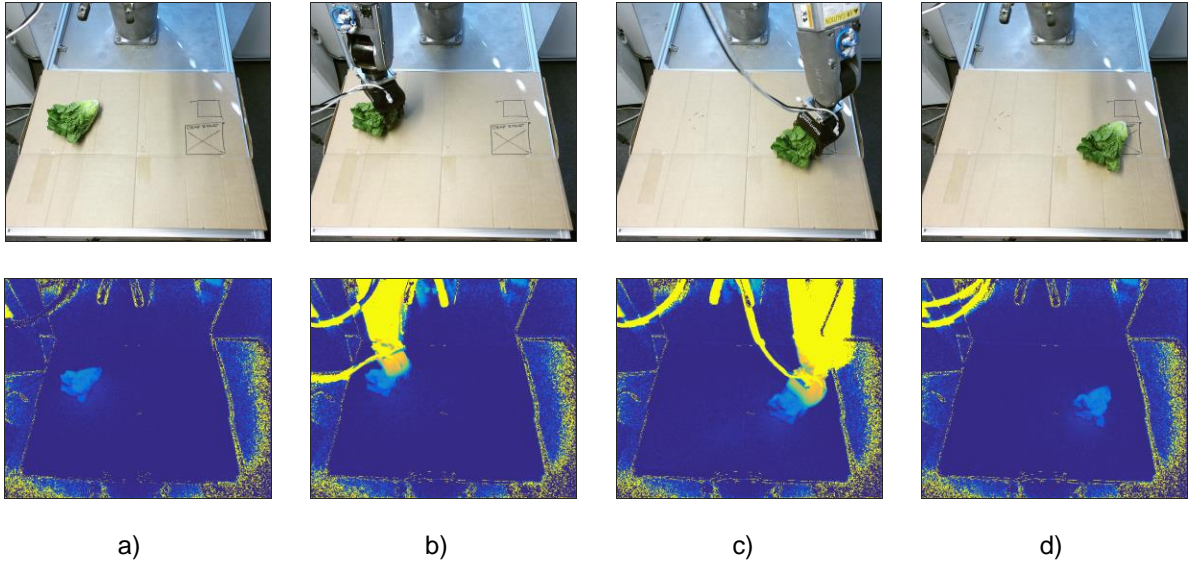

Figure 5. The gripping sequence shown in RGB (top row) and depth (bottom row) images based on our trained *LfD* learning policy, where in a) an initial image is acquired and the visual state of the lettuce is computed, b) the robot places a grasp on the lettuce according to the action derived from the visual state, c) the robot moves and releases the lettuce to a predefined target point, d) the robot moves out of the way, enabling visual confirmation of whether the grasping sequence succeeded.

### C. State and action spaces

The state space consists of vectors **s** describing visual features of the lettuce. These visual features are illustrated in Fig. 4. All features are extracted from the depth image of the Kinect v2 camera. After subtracting a reference image from the depth image, we obtain a height image that is approximately zero for the surface on which the lettuce is placed, and greater than zero for the lettuce. The lettuce is found by thresholding the height image, resulting in a binary mask. The first set of features is the centroid in the binary mask, which we denote by *a* in Fig. 4, for which we compute the coordinates $(x_a, y_a, z_a)$ in the robot reference frame. The value of the height image at the centroid is denoted by $h_a$. The major and minor axis of the binary mask gives us the angle $\theta$ from the horizontal axis in the image, and the length of these axes give us the lettuce length $l_a$ and width $w_a$, respectively. To enable convex regression of $\theta$, we include both components of the unit-length direction vector in the state space vector. To estimate the leaf end, leaf spread and height of the lettuce, we compute two additional points *b* and *c*, as seen in Fig. 4, halfway between the centroid *a* and the respective end points of the major axis. The heights $h_b$ and $h_c$ are computed at these halfway points, as well as the widths $w_b$ and $w_c$. Concatenating all these features, gives us the state vector

$$\boldsymbol{s} = [x_a \ y_a \ z_a \ h_a \ l_a \ w_a \ \cos\theta \ \sin\theta \ h_b \ w_b \ h_c \ w_c].$$

The action space consists of vectors $\boldsymbol{a}$ describing the lettuce grasp to be performed by the robot and its gripper. The robot end-effector pose is described by its position $\boldsymbol{p} = [p_x \ p_y \ p_z]$, and its orientation described by two orthonormal vectors $\boldsymbol{d}^1$ and $\boldsymbol{d}^2$ and an additional vector $\boldsymbol{d}^3$, each vector being of the form $\boldsymbol{d}^k = [d_x^k \ d_y^k \ d_z^k]$, $k \in \{1, 2, 3\}$. Although two orthonormal vectors suffice to uniquely describe an orientation in $\mathbb{R}^3$, the three vectors provide redundancy for a more robust regression when two ambiguous grips exist. The gripper configuration is described by a finger figure $f$ and preshape $pre$. The gripper continues to close its grip until certain significant pressure thresholds $spt_1$, $spt_2$ and $spt_3$ are obtained in the tactile sensors in each of the three respective fingers. Each significant pressure threshold for one finger represents the mean of all tactile sensor values that are over 70% of the maximum value. When the significant pressure threshold is met in *one* of its fingers, the gripper stops closing. Concatenating these parameters, gives us the following action vector:

$$\boldsymbol{a} = [\boldsymbol{p} \ \boldsymbol{d}^1 \ \boldsymbol{d}^2 \ \boldsymbol{d}^3 \ f \ pre \ spt_1 \ spt_2 \ spt_3].$$

### VI. RESULTS AND DISCUSSION

In this section, we present the experimental results of our *LfD* learning policy by a quantitative and qualitative assessment. The entire system, including robot arm, gripper, depth camera and learning algorithm, was tested in a teaching phase (Fig. 3) and an execution phase (Fig. 5). In the teaching phase, a user provided recorded demonstrations of the form described in Section V, which were then used to derive a learning policy as described in Section IV. In the execution phase, the learning policy was validated to teaching a robot to grasp lettuces based on visual and tactile information.

The teaching phase consisted of 525 demonstrations (Fig. 3a, b, c), where a human operator used the STEM controllers to tele-operate the robot to place the gripper around the lettuce.

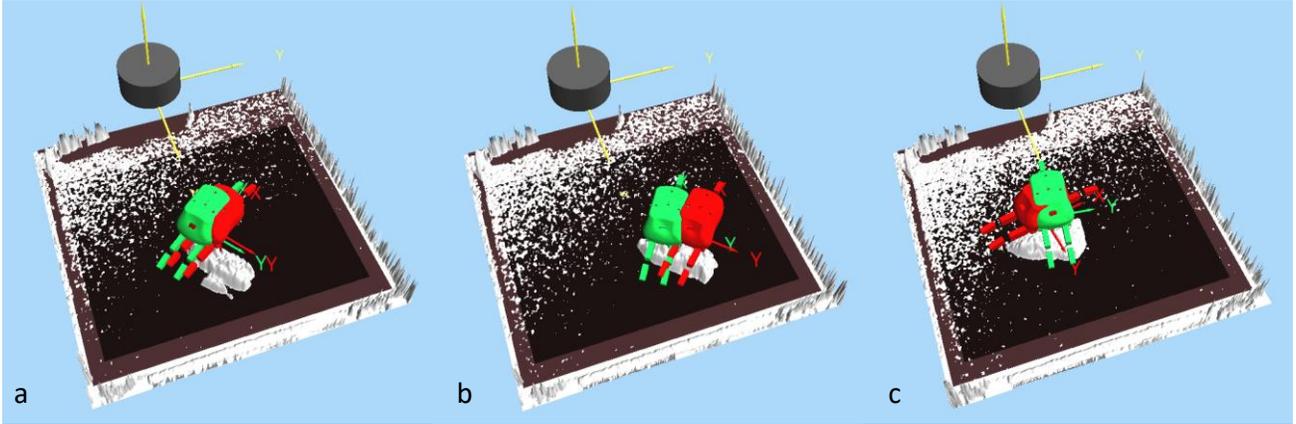

Figure 6. Visualization of the gripper pose during demonstration (green) and after training during autonomous grasping based on a learned policy (red) to qualitatively highlight some of the challenges in the estimation of the correct pose. While there is a general good estimation of the gripper position for robotic grasping (a, b), we also show an example of failure to predict correct orientation of the gripper (c).

A batch of 20 lettuce with variation in size and shape was used, placed in random positions and orientations on the grasping area (Fig 3). The teaching phase, with demonstrations from human teacher, followed the steps illustrated in Fig. 3, while the execution phase (after the learned policy) is illustrated in Fig. 5. In step a) (Fig. 5), color and depth images were acquired with the robot positioned in a start pose out of view of the camera. The colored (RGB) depth image was used to extract visual states constituting the state space vector in both the teaching and execution phases. Although colour was not used, it is useful information for visual tracking in food production lines with moving objects. In step b), the robot moved based on a learnt policy in the execution phase to assume the gripper pose for grasping, while in step c), the lettuce was grasped, lifted (Fig. 8), and moved to a target drop point, whereupon in step d) the gripper released the lettuces and the robot returned to its starting position. Images were recorded in steps c) and d) for verification of success both in the teaching and execution phase after learning. In our experiments, 497 of the total 525 demonstrations from human teacher were successful, resulting in 28 inconsistent examples that were automatically removed by the *LfD* algorithm. After analysis, the inconsistent examples consisted on either bad orientation of the gripper when attempting to grasp, or bad grip during the demonstration phase from the teacher. In Fig. 6 is shown a visualization of the gripper pose in the demonstration-training phase (green gripper) versus execution phase (red gripper), highlighting the nature of the inconsistent demonstrations consisting on orientation. In Fig. 7, we show the average success rate on the autonomous grasping of lettuces on the test set, resulting in an average accuracy of 75%. Here, 14 lettuces of different sizes and shapes constituted the test set, and each lettuce was grasped 5 times in random positions on the grasping area. The failures were associated to either bad prediction in orientation of the gripper or insufficient grasping force from fingers resulting in a bad grip and consequent slip. Lettuces are characterized by both high biological variation in size and shape, in addition their compliancy has a high variation from the leaves side towards the root. Some of the grasp failures were associated to correctly predicted finger pressures but resulting in a bad grip due to an incorrect orientation of the gripper, positioned more on the leaf side of the lettuce.

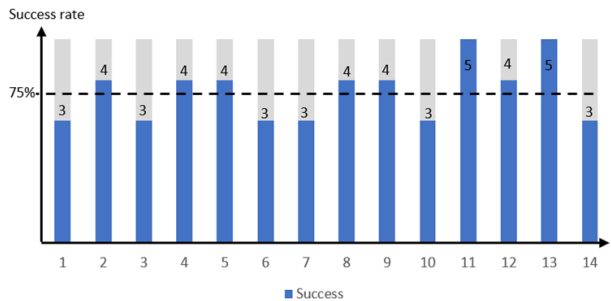

Figure 7. Average success rate on autonomous grasping on the test set after learning policy. A total of 14 lettuces of different size and shape were used to test the learned policy, and each lettuce was grasped 5 times from random positions. The overall size for the testing set was 70 grasps resulting in 75% average success rate. The number on the bars is the number of successful grasps out of 5 attempts for each lettuce randomly placed on the grasping area.

TABLE I: Overall comparison of the *LfD* learning policy that automatically discards the inconsistent demonstrations of the human teacher. It is seen that automatic discarding of the inconsistent examples results in higher accuracy given here by the average regression coefficient and standard deviation for gripper position ($p_x$, $p_y$, $p_z$), orientation ($R_x$, $R_y$, $R_z$), and fingers' significant pressure/tactile threshold ($spt_1$, $spt_2$, $spt_3$).

| *LfD* policy | Correlation $R^2$ | STD |
|---|---|---|
| Consistent only | 0.60 | 0.31 |
| Including inconsistent | 0.545 | 0.33 |

In Table I, we show the accuracy in the learned policy by comparing the accuracy between including the inconsistent human teacher's demonstrations versus automatic discarding of inconsistent demonstrations. The experimental results show a higher accuracy in the value of the coefficient $R^2$, related to position ($p_x$, $p_y$, $p_z$), orientation ($R_x$, $R_y$, $R_z$), and tactile values ($spt_1$, $spt_2$, $spt_3$), generated by the learned policy when the inconsistent demonstrations were automatically discarded.

Explicit enforcing of few support vectors is also seen as a way to assist in acquiring a more robust learning policy. The focus on more cases, where the human teacher has a greater challenge in providing accurate demonstrations so that we may explore the potential of our method for discovering the teacher's intended policy, may be a natural extension of the current study. Examples of such challenging demonstration tasks where the human teacher may have challenges to provide good demonstrations are: grasping of tiny fillet portion-bits of fish fillets which are challenging to grasp even with human hand due to the texture and stickiness towards the surface on which they are placed; cutting of ham carcasses with the purpose of teaching the robot to accomplish the same. Cutting of ham carcasses is a very challenging and complex task requiring human experience and involving both visual and tactile perception and high degree of dexterity.

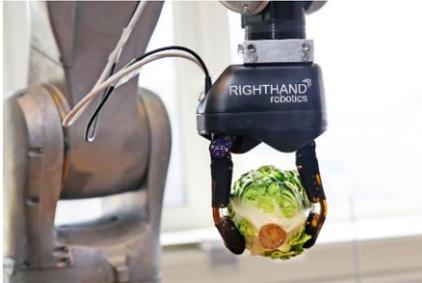

Figure 8. A snapshot of a close-up image of an autonomous robotic grasping of lettuce based on *LfD* learning policy.

## VII. CONCLUSIONS

In this paper, we proposed a robust learning policy based on Learning from Demonstration (*LfD*) addressing the problem of learning when presented with inconsistent human demonstrations. We presented an *LfD* approach to the automatic rejection of inconsistent demonstrations by human teachers. The approach is validated by successfully teaching the robot to grasp fragile and compliant food objects such as lettuces. It utilises the merging of RGB-D images and tactile data in order to estimate the pose of the gripper, the gripper's finger configuration, and the forces exerted on the object in order to achieve successful grasping. The robust policy learning method presented here will enable the learner (robot) to act more consistently and with less variance than the teacher (human). Such robust learning algorithm also alleviates the burden of accuracy on the human teacher. The proposed approach has a vast range of potential applications in the ocean space, agriculture, and food industries, where the manual nature of processes leads to high variation in the way skilled operators perform complex processing and handling tasks.

## ACKNOWLEDGMENT

The work is supported by The Research Council of Norway in iProcess – 255596 project.